\documentclass[conference,a4paper]{IEEEtran}
\usepackage{cite}
\usepackage{amsmath,amssymb,amsfonts}
\usepackage{algorithmic}
\usepackage{graphicx}
\usepackage{textcomp}
\usepackage{xcolor}
\usepackage{booktabs}
\usepackage{multirow}
\usepackage{makecell}
\usepackage{subcaption}
\usepackage{url}
\usepackage{dblfloatfix}
\usepackage{tikz}
\usepackage{listings}
\usepackage{breakurl}
\lstdefinestyle{promptstyle}{
  basicstyle=\ttfamily\scriptsize,
  breaklines=true,
  breakindent=0pt,
  columns=fullflexible,
  frame=single,
  framesep=3pt,
  rulecolor=\color{black!35},
  backgroundcolor=\color{black!3},
  xleftmargin=2pt,
  xrightmargin=2pt,
  aboveskip=2pt,
  belowskip=2pt,
  keepspaces=true,
  showstringspaces=false,
  literate={->}{{$\rightarrow$}}2,
}

\usepackage{eso-pic}
\AddToShipoutPictureBG*{%
  \AtPageLowerLeft{%
    \put(\LenToUnit{0.08\paperwidth},\LenToUnit{0.03\paperheight}){%
      \parbox{0.84\paperwidth}{\scriptsize
        \copyright\ 2026 IEEE. Personal use of this material is permitted.
        Permission from IEEE must be obtained for all other uses, in any current
        or future media, including reprinting/republishing this material for
        advertising or promotional purposes, creating new collective works, for
        resale or redistribution to servers or lists, or reuse of any
        copyrighted component of this work in other works.}
        }
    }
}

\def\BibTeX{{\rm B\kern-.05em{\sc i\kern-.025em b}\kern-.08em
    T\kern-.1667em\lower.7ex\hbox{E}\kern-.125emX}}
\begin{document}
\setcounter{topnumber}{4}
\setcounter{dbltopnumber}{4}
\renewcommand{\topfraction}{0.95}
\renewcommand{\dbltopfraction}{0.95}
\renewcommand{\textfraction}{0.05}
\renewcommand{\floatpagefraction}{0.8}
\renewcommand{\dblfloatpagefraction}{0.8}

\setlength{\abovecaptionskip}{4pt}
\setlength{\belowcaptionskip}{0pt}
\setlength{\textfloatsep}{8pt plus 2pt minus 2pt}
\setlength{\intextsep}{8pt plus 2pt minus 2pt}
\setlength{\floatsep}{8pt plus 2pt minus 2pt}

\title{Design and Evaluation of LLM Chaining-Based Task Planning
       for General Purpose Service Robots}
\author{
\IEEEauthorblockN{Lucas Da Mota Bruno}
\IEEEauthorblockA{\textit{Grad. School of Science and Engineering} \\
\textit{Soka University}\\
Tokyo, Japan \\
lucasmotabr@icloud.com}
\and
\IEEEauthorblockN{Jiahao Sim}
\IEEEauthorblockA{\textit{Grad. School of Science and Engineering} \\
\textit{Soka University}\\
Tokyo, Japan \\
simjiahao@ieee.org}
\and
\IEEEauthorblockN{Yoshinobu Hagiwara}
\IEEEauthorblockA{\textit{Faculty of Science and Engineering} \\
\textit{Soka University}\\
Tokyo, Japan \\
hagiwara@soka.ac.jp}
}
\maketitle

\begin{abstract}
General Purpose Service Robot (GPSR) tasks, as defined in the
RoboCup@Home benchmark~\cite{wisspeintner2009}, require robots to
interpret diverse natural language commands and generate multi-step
action sequences in real home environments.
Conventional Single Prompt (SP) approaches suffer from context bloat
and the ``Lost in the Middle'' phenomenon, leading to unreliable task
planning.
We propose an LLM chaining architecture that separates instruction
classification and action generation into two specialized stages,
reducing per-inference prompt length by approximately 45\% while
improving planning consistency.
We evaluate our method using 100 randomly generated GPSR commands
across three language models spanning local open-source and frontier
cloud deployment contexts.
Results show consistent planning improvements over SP across all
models, with gains of up to +37 percentage points on local models.
Further, real-robot execution experiments on the Toyota Human Support
Robot (HSR) reveal that planning success alone does not guarantee
task completion, with 6 of 10 tasks completing successfully and
execution-layer failures identified as the primary remaining
bottleneck.
\end{abstract}
\begin{IEEEkeywords}
service robots, task planning, large language models,
GPSR, RoboCup@Home
\end{IEEEkeywords}

\section{Introduction}
General Purpose Service Robot (GPSR) tasks~\cite{wisspeintner2009}
represent one of the most demanding benchmarks in the RoboCup@Home
competition, requiring the interpretation of unconstrained natural
language commands and the generation of precise, multi-step action
sequences in real home environments.
As large language models (LLMs) have demonstrated strong general
reasoning capabilities for robot control, prior
service robot systems have adopted a Single Prompt (SP) method that
aggregates all task knowledge --- function definitions, environment
constraints, and execution examples --- into a single monolithic
prompt~\cite{hasegawa2025,shirasaka2023,liang2023}.

However, SP approaches face two structural problems in GPSR contexts:
prompts rapidly grow with task diversity, causing inference delays
and out-of-memory failures in local deployments; and LLMs exhibit
the ``Lost in the Middle'' phenomenon~\cite{liu2024}, where
mid-context information is systematically under-attended, leading
to constraint violations and hallucinated parameters.
This matters most for local deployment.
Cloud APIs incur per-request cost that accumulates when a robot
repeats long-horizon tasks, network latency degrades responsiveness
in real-time control, and competition venues and home environments
frequently have unreliable connectivity --- so dependence on a cloud
API is itself a reliability risk.
Local models avoid these constraints structurally, but their smaller
scale makes stable generation under SP difficult.

We propose an LLM chaining architecture decomposing GPSR task
planning into two sequential stages: an Instruction Classifier
routing commands to one of 28 task categories, and an Action
Generator producing step-by-step sequences using a task-specific
action schema.
This work extends our earlier study~\cite{damota2025}, which
validated the chaining structure on 15 tasks and three models, to a
100-command benchmark including a frontier cloud model.
We make the following contributions:
\begin{itemize}
    \item A two-stage LLM chaining architecture reducing
    per-inference context length by approximately 45\%.
    \item A large-scale benchmark evaluating 100 GPSR commands
    across three language models spanning local and frontier
    deployment contexts.
    \item A real-robot execution analysis on the HSR platform
    identifying execution-layer failures as the primary bottleneck.
\end{itemize}

\begin{figure*}[!t]
    \centering
    \includegraphics[width=0.85\textwidth]{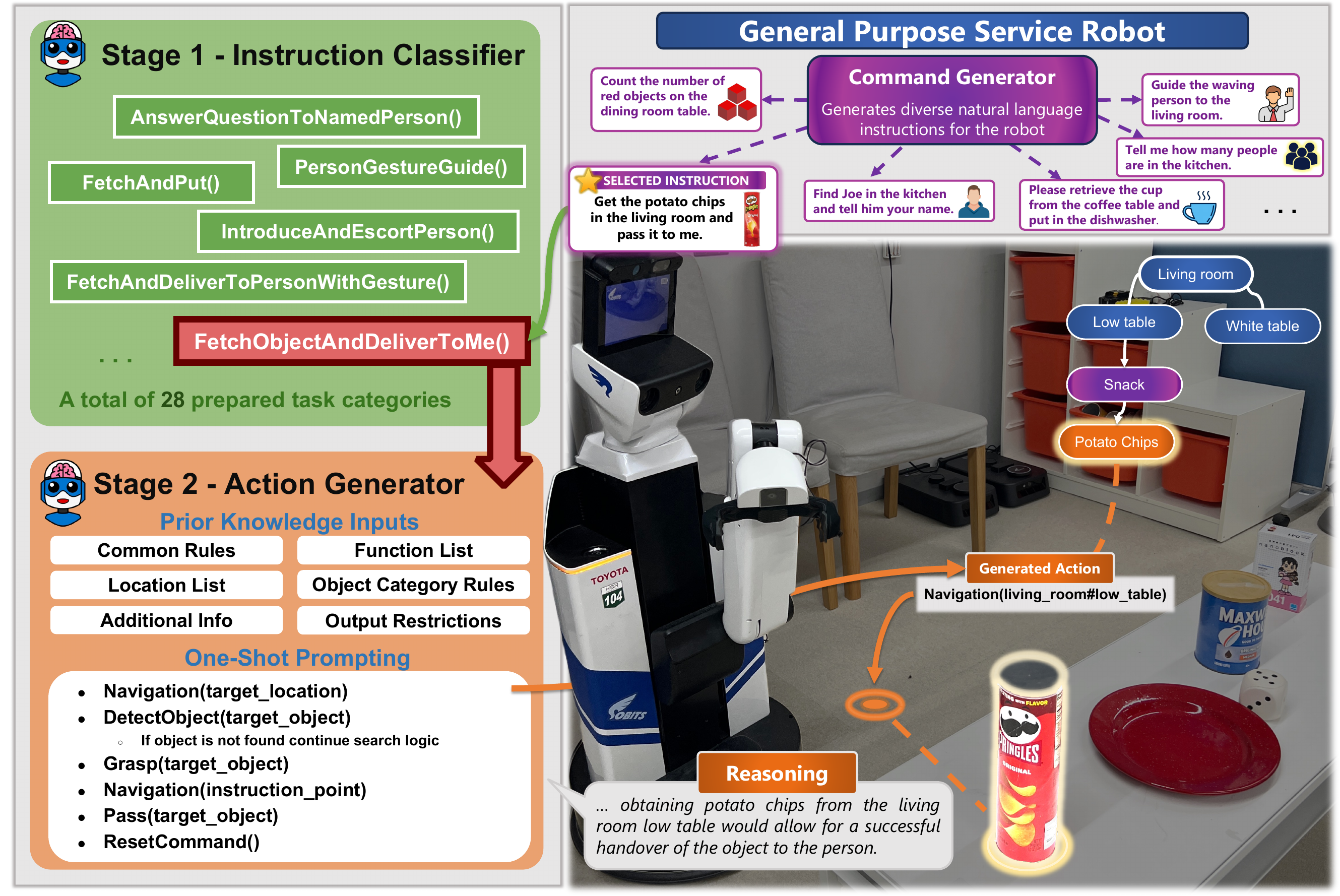}
    \caption{Overview of the proposed LLM chaining framework
    deployed on the HSR platform. Left: the two-stage pipeline
    (Instruction Classifier~$\rightarrow$~Action Generator).
    Right: real-robot execution with semantic map overlay and
    generated navigation action.}
    \label{fig:system}
    \vspace{-5pt}
\end{figure*}

\section{Related Work}
LLM-based task planning for service robots has largely followed the
single-prompt paradigm.
Liang et al.~\cite{liang2023} generate executable policy code
directly from a single prompt containing the full API surface.
Shirasaka et al.~\cite{shirasaka2023} construct a promptable GPSR
system with foundation models and add a self-recovery mechanism that
re-plans after a failure is detected; recovery is triggered
downstream of planning, so instability in the initial plan is
handled reactively rather than prevented.
Hasegawa et al.~\cite{hasegawa2025} incorporate spatial concepts
into the prompt to ground action planning in learned place
representations, improving the plausibility of generated plans but
retaining a single monolithic prompt whose length grows with the
size of the environment model.
Liu et al.~\cite{liu2024} show that language models systematically
under-attend to information in the middle of long contexts, which
bounds how far the single-prompt approach can scale as task
diversity increases.
Our work differs in addressing the planning stage itself: rather
than recovering from failed plans or enriching a single prompt, we
decompose planning into two shorter, specialized inferences so that
no single prompt carries the full task knowledge.

\section{Proposed Method}
\subsection{Architecture Overview}
Fig.~\ref{fig:system} illustrates the proposed framework on the HSR;
a natural language command passes through two sequential LLM stages
before actions are dispatched to the robot.
The two stages are described in turn below, each alongside an
excerpt of the prompt that drives it.

\subsection{Stage 1: Instruction Classification}
The Instruction Classifier outputs exactly one of 28 predefined GPSR
task categories from the raw command, using few-shot prompting with
boundary disambiguation examples to handle semantically similar
categories (e.g., delivery to a named person vs.\ to a person
identified by visual feature).
Fig.~\ref{fig:prompt1} shows the structure of this prompt: a fixed
category list, followed by explicit disambiguation rules for the
boundaries that proved hardest in practice, and an output
restriction that forbids any text other than the category name.
The 28 categories were determined through systematic analysis of the
RoboCup@Home GPSR command space: by examining commands generated
by the Command Generator~\cite{commandgenerator2025}, we identified
the minimal set of task types sufficient to cover the full range
of required robot behaviors, arriving at 28 categories through
iterative refinement.

\begin{figure}[!t]
\begin{lstlisting}[style=promptstyle]
--- STAGE 1: INSTRUCTION CLASSIFIER ---
You are a classification-only AI. Output exactly one
group name from GROUP_LIST. Never explain.

GROUP_LIST (28 categories):
- FetchObjectAndDeliverToMe: fetch an object and
  bring it to the host at instruction_point.
- FetchAndDeliverToPersonWithFeature: deliver to a
  person identified by a visual feature.
- FetchAndDeliverToPersonWithGesture: deliver to a
  person performing a gesture.
  ... (25 further categories)

Disambiguation rules:
- Deliver by gesture (NOT name)  -> WithGesture
- Deliver by feature (NOT gesture) -> WithFeature
- Deliver to the host -> FetchObjectAndDeliverToMe

Output only the group name. No extra text.
\end{lstlisting}
\caption{Condensed excerpt of the Stage~1 classifier prompt.
Ellipses mark omitted categories.}
\label{fig:prompt1}
\end{figure}

\subsection{Stage 2: Action Generation}
The Action Generator loads the task-specific action schema for
the classified category, generating executable robot function calls
one step at a time:
observe $\rightarrow$ generate $\rightarrow$ execute $\rightarrow$ feedback.
Fig.~\ref{fig:prompt2} shows the corresponding prompt for one
category.
Only the schema for the classified category is loaded, so the
generator never sees the action plans of the other 27; the fixed
function list and location list bound what it may emit, and the
ordering constraints specific to that category are stated inline
rather than buried among rules for unrelated tasks.
The Stage~2 context length is reduced from approximately 1,802 tokens
(SP) to 992 tokens on average per inference step ($\approx$45\% reduction);
the Stage~1 classifier runs in a separate context of approximately
1,168 tokens, so neither stage accumulates the other's prompt.
Beyond efficiency, this is mechanistically important since smaller
local models begin ignoring buried mid-context instructions as SP
length grows, causing complete output collapse.
Output stability is further enforced through stop-token injection
at function call boundaries, preventing multi-step hallucination.

\begin{figure}[!t]
\begin{lstlisting}[style=promptstyle]
--- STAGE 2: ACTION GENERATOR ---
Schema loaded: FetchObjectAndDeliverToMe

Output one function per step, "Function(arg)" format.
function_list: Navigation(arg), DetectObject(arg),
  Grasp(arg), Pass(arg), MoveToPerson(arg),
  DetectPerson(), Speak(arg), Listen(arg),
  ResetCommand(), ...
location_list: instruction_point,
  kitchen#side_table, kitchen#dining_table,
  kitchen#cabinet, ...

ACTION PLAN:
  F1: Navigation([object_location])
  F2: for each sub_location in room:
        Navigation([sub_location])
        DetectObject([target_object])
        if found -> break
  F3: Grasp([target_object])
  F4: Navigation(instruction_point)
  F5: Pass([target_object])
  F6: ResetCommand()

Constraints: one function per output; never emit a
bare room name; only locations in location_list.
\end{lstlisting}
\caption{Condensed excerpt of the Stage~2 generator prompt for one
category. A different action plan is loaded for each of the 28
categories; the function and location lists are shared.}
\label{fig:prompt2}
\end{figure}

\begin{figure}[!t]
\centering
\resizebox{0.96\columnwidth}{!}{%
\begin{tikzpicture}[
  x=1cm, y=1cm,
  every node/.style={font=\small},
  furn/.style={draw, thick, fill=blue!4},
  room/.style={font=\small\bfseries},
  wall/.style={line width=1.1pt},
]
\draw[wall] (0,10.3) -- (10.0,10.3);
\draw[wall] (0,0)    -- (0,10.3);
\draw[wall] (10.0,0) -- (10.0,10.3);
\draw[wall] (0,0)    -- (1.38,0);
\draw[wall] (3.04,0) -- (6.91,0);
\draw[wall] (8.39,0) -- (10.0,0);

\draw[wall] (4.89,10.3) -- (4.89,8.48);
\draw[wall] (4.89,7.10) -- (4.89,3.69);
\draw[wall] (4.89,2.49) -- (4.89,0);
\draw[wall] (1.38,4.50) -- (4.89,4.50);
\draw[wall] (4.93,4.50) -- (6.27,4.50);
\draw[wall] (8.39,4.50) -- (10.0,4.50);

\node[room] at (2.58,7.56) {Laundry Room};
\node[room] at (7.47,7.56) {Kitchen};
\node[room] at (2.77,1.80) {Living Room};
\node[room] at (5.97,3.99) {Bedroom};

\node[furn, minimum width=1.06cm, minimum height=0.89cm]
     at (2.46,9.53) {11};
\node[furn, minimum width=3.37cm, minimum height=1.57cm]
     at (3.08,5.35) {10};

\node[furn, minimum width=1.57cm, minimum height=0.51cm]
     at (6.13,9.71) {9};
\node[furn, minimum width=0.60cm, minimum height=0.46cm]
     at (7.90,9.72) {8};
\node[furn, rotate=45, minimum width=0.62cm, minimum height=0.62cm]
     at (9.45,9.56) {7};
\node[furn, minimum width=1.15cm, minimum height=1.86cm]
     at (9.28,7.48) {6};
\node[furn, minimum width=0.75cm, minimum height=0.77cm]
     at (5.50,5.06) {4};
\node[furn, minimum width=0.55cm, minimum height=0.74cm]
     at (9.48,5.12) {5};

\node[furn, minimum width=0.64cm, minimum height=0.50cm]
     at (2.23,4.01) {14};
\node[furn, minimum width=2.00cm, minimum height=0.60cm]
     at (3.76,3.99) {13};
\node[furn, minimum width=1.13cm, minimum height=2.09cm]
     at (0.63,1.10) {12};

\node[furn, minimum width=1.43cm, minimum height=2.14cm]
     at (5.81,1.16) {1};
\node[furn, minimum width=0.76cm, minimum height=0.97cm]
     at (9.49,0.67) {2};
\node[furn, circle, minimum size=0.60cm] at (9.36,3.78) {3};

\fill[blue!55!black] (7.82,1.89) circle (5pt);
\draw[->, line width=2.2pt, blue!55!black] (7.82,1.89) -- (7.82,2.95);
\node[anchor=west, align=left] at (8.15,2.05) {Instruction\\Point};

\node at (2.2,-0.55) {\bfseries Exit};
\node at (7.6,-0.55) {\bfseries Entrance};
\end{tikzpicture}}

\vspace{4pt}
{\footnotesize
\begin{tabular}{@{}p{0.44\columnwidth}p{0.44\columnwidth}@{}}
1. Bed & 8. Microwave \\
2. Bookcase & 9. Cabinet \\
3. Coat rack & 10. Table \\
4. Side Table & 11. Washing Machine \\
5. Trash Bin & 12. Coffee Table \\
6. Dining Table & 13. Couch \\
7. Dishwasher & 14. Chair \\
\end{tabular}}

\caption{GPSR arena layout used in the real-robot experiments:
four rooms connected by open doorways, 14 furniture locations, and
a fixed instruction point. Layout follows the RoboCup@Home Japan
GPSR definition~\cite{objectlist2026}.}
\label{fig:arena}
\end{figure}
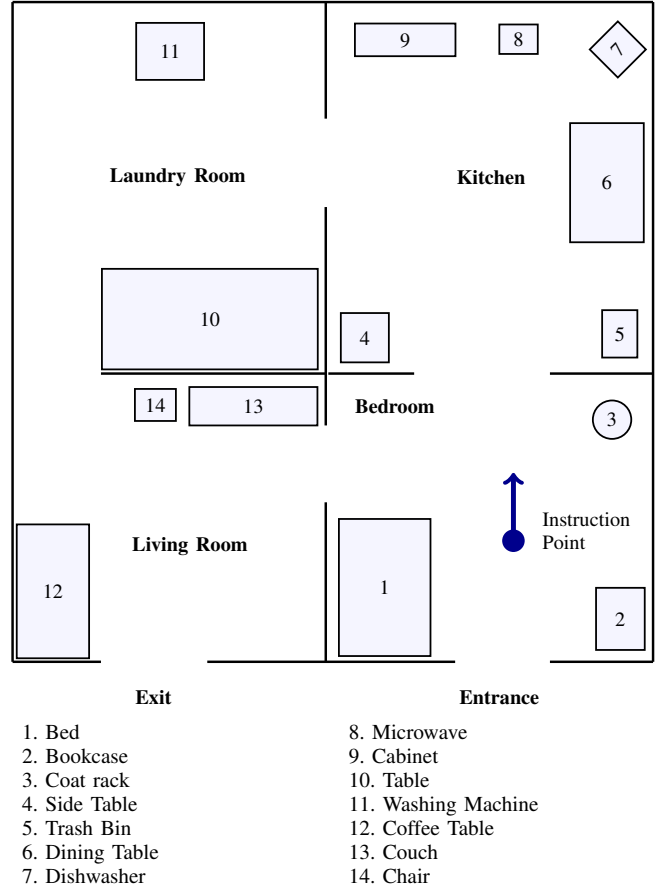

\section{Experiments}
\subsection{Large-Scale Planning Benchmark}
\subsubsection{Setup}
We evaluated 100 GPSR commands randomly generated by the RoboCup@Home
Japan Command Generator~\cite{commandgenerator2025} across all 28
task categories, with no overlap with the few-shot prompting examples.
Objects referenced by the commands are drawn from the RoboCup@Home
Japan 2026 DSPL object list~\cite{objectlist2026}, which defines 43
items across six categories (food, snack, fruit, drink, kitchen item,
and task item), combining YCB objects with commercially available
Japanese products.
Three models were evaluated: Qwen2.5-14B and Cogito-14B (local
open-source) and Claude Sonnet~4.6 (frontier cloud), each under
both SP and chaining.
Local models were run on a laptop with an Intel Core~i7-12700H
processor and an NVIDIA RTX~3080~Ti (16~GB) GPU; the frontier model
was accessed via API.
In the chaining condition, the task category was supplied from the
ground-truth label rather than predicted by the Stage~1 classifier,
isolating action generation from classification error.

\subsubsection{Metrics}
\emph{Success rate (SR)} counts a task as successful if the complete
action sequence is generated without output breakdown (format
collapse, undefined function calls, hallucinated arguments, or
sequence order violations), further validated through manual review
to exclude semantically incomplete plans that passed automated
checks.
\emph{AUC-E} integrates the cumulative success
rate over an increasing step budget, jointly rewarding early success
and resistance to early breakdown.
\emph{Steps to end} is the mean number of steps generated before a
run terminates in success or failure, and measures how long a method
sustains a coherent plan.

\begin{table}[!t]
\caption{Planning Success Rate (\%): SP vs.\ Chaining}
\label{tab:sr}
\centering
\setlength{\tabcolsep}{6pt}
\renewcommand{\arraystretch}{1.25}
\normalsize
\begin{tabular}{llccc}
\toprule
\textbf{Model} & \textbf{Type} & \textbf{SP} & \textbf{Chaining}
               & \textbf{$\Delta$ (pp)} \\
\midrule
Qwen2.5-14B       & Local & 7  & \textbf{33} & +26 \\
Cogito-14B        & Local & 17 & \textbf{54} & +37 \\
Claude Sonnet~4.6 & Cloud & 81 & \textbf{88} & +7  \\
\bottomrule
\end{tabular}
\end{table}

\begin{figure*}[!t]
    \centering
    \begin{minipage}[t]{0.485\textwidth}
        \centering
        \includegraphics[width=\linewidth]{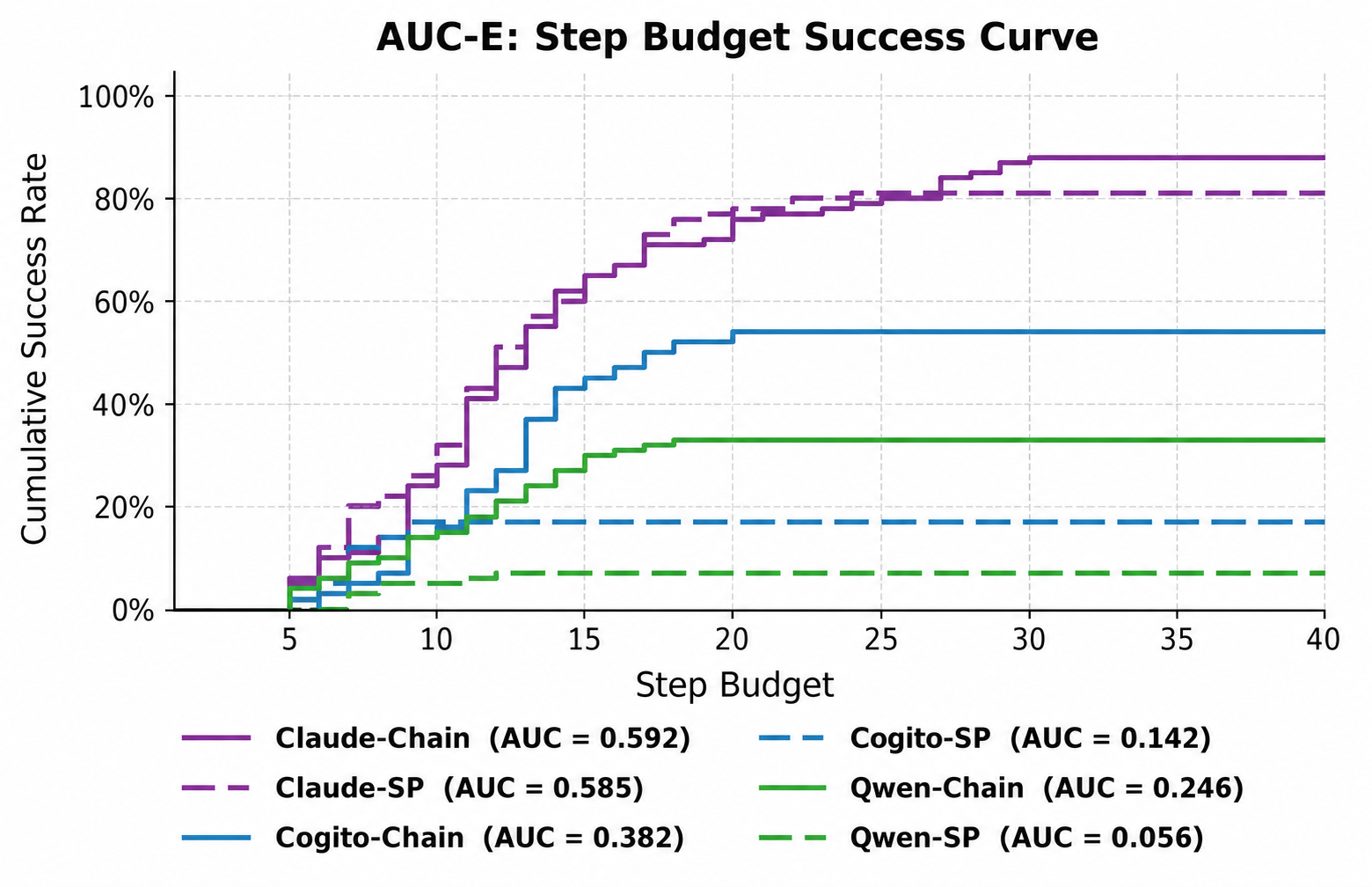}
        \caption{AUC-E (Area Under the Execution-efficiency Curve):
        chaining maintains higher cumulative success rates as step
        budget increases, with large gains on local models.}
        \label{fig:auce}
    \end{minipage}
    \hfill
    \begin{minipage}[t]{0.485\textwidth}
        \centering
        \includegraphics[width=\linewidth]{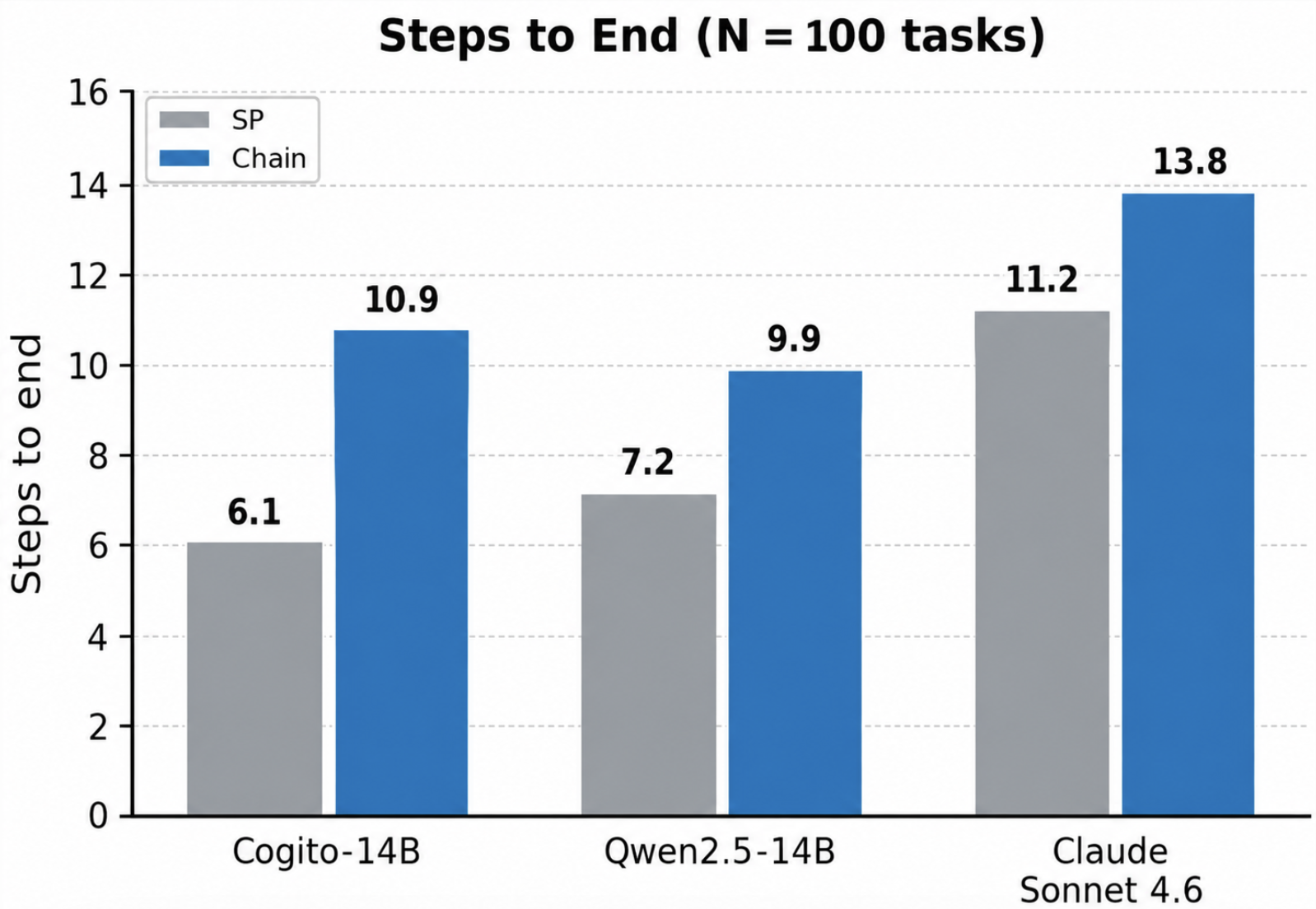}
        \caption{Steps to end (mean over 100 tasks): chaining sustains
        longer action sequences before terminating, on both local and
        frontier models.}
        \label{fig:steps}
    \end{minipage}
    \vspace{-4pt}
\end{figure*}

\subsubsection{Results}
Table~\ref{tab:sr} shows chaining consistently outperforms SP across
all models, with gains from +7~pp (frontier) to +37~pp (local):
Cogito-14B rises from 17\% to 54\% and Qwen2.5-14B from 7\% to 33\%.
The same $+26$ to $+37$~pp margin was observed at the 15-task scale
in our earlier study~\cite{damota2025}, indicating the effect is
stable as the benchmark grows.
Fig.~\ref{fig:auce} shows the AUC-E curves; local gains are large
(Cogito: +0.240, Qwen: +0.190), while frontier values are nearly
equal (0.592 vs.\ 0.585).
Manual review revealed that SP successes for the frontier model
frequently omit schema-required intermediate steps, passing format
checks while producing incomplete sequences.

Fig.~\ref{fig:steps} reports steps to end.
Chaining sustains longer action sequences on every model
(Cogito: 6.1$\rightarrow$10.9, Qwen: 7.2$\rightarrow$9.9,
Claude: 11.2$\rightarrow$13.8).
Because GPSR plans require a sufficient number of steps to complete,
this indicates that SP runs terminate early rather than producing
wrong but complete plans.
Consistent with this, SP failures fell into three recurring modes:
omission of required function arguments; violation of the ordering
constraints encoded in the schema; and mid-sequence output collapse,
in which generation degenerates into free-form text before the plan
terminates.
The last mode is consistent with the ``Lost in the Middle''
effect~\cite{liu2024}: the SP prompt is roughly twice as long
and mixes constraints for many task types, so the ordering rules
relevant to the current command are more likely to be
under-attended than in the short, single-category context that
chaining supplies.
Table~\ref{tab:example} contrasts the two methods on a single
command, illustrating an ordering violation.

\begin{table}[!b]
\caption{Planning Example: Chaining vs.\ SP on the Same Command}
\label{tab:example}
\centering
\footnotesize
\begin{tabular}{@{}p{0.13\columnwidth}p{0.79\columnwidth}@{}}
\toprule
\multicolumn{2}{@{}l}{\textbf{Command:} ``Take the cola on the table
and pass it to me.''} \\
\midrule
\multicolumn{2}{@{}l}{\textbf{Chaining (success)}} \\
Stage 1 & \texttt{FetchObjectAndDeliverToMe} \\
Stage 2 & \texttt{Navigation(dining\_table)} \\
        & \texttt{DetectObject(cola)} \\
        & \texttt{Grasp(cola)} \\
        & \texttt{Navigation(instruction\_point)} \\
        & \texttt{Pass()} \\
\midrule
\multicolumn{2}{@{}l}{\textbf{SP (failure --- ordering violation)}} \\
Output  & \texttt{Navigation(dining\_table)} \\
        & \texttt{DetectObject(cola)} \\
        & \texttt{Grasp(cola)} \\
        & \texttt{Pass()} \\
Cause   & \texttt{Pass()} issued without first returning to
          \texttt{instruction\_point}; the object is released at the
          table rather than delivered to the host. \\
\bottomrule
\end{tabular}
\end{table}

\subsection{Real-Robot Execution Analysis}
\subsubsection{Setup}
We executed 10 GPSR tasks on the HSR~\cite{yamamoto2019hsr} in a
laboratory reproduction of the RoboCup@Home Japan GPSR arena
(Fig.~\ref{fig:arena}), with four rooms, 14 furniture locations, and
a fixed instruction point.
These locations plus per-room people-search targets form the
\texttt{location\_list} supplied to the Action Generator.
Failures were classified as Recognition, Navigation, or Manipulation.
Object perception uses SAM~3~\cite{sam3} directly as a
text-promptable detector, with no separate grounding model; person
detection and gesture recognition run on a parallel YOLO11n-pose
branch.

\subsubsection{Results}
\begin{figure*}[!t]
    \centering
    \setlength{\fboxsep}{0pt}%
    \setlength{\fboxrule}{0.4pt}%
    \begin{subfigure}{0.327\textwidth}
        \centering
        \fbox{\includegraphics[height=3.2cm]{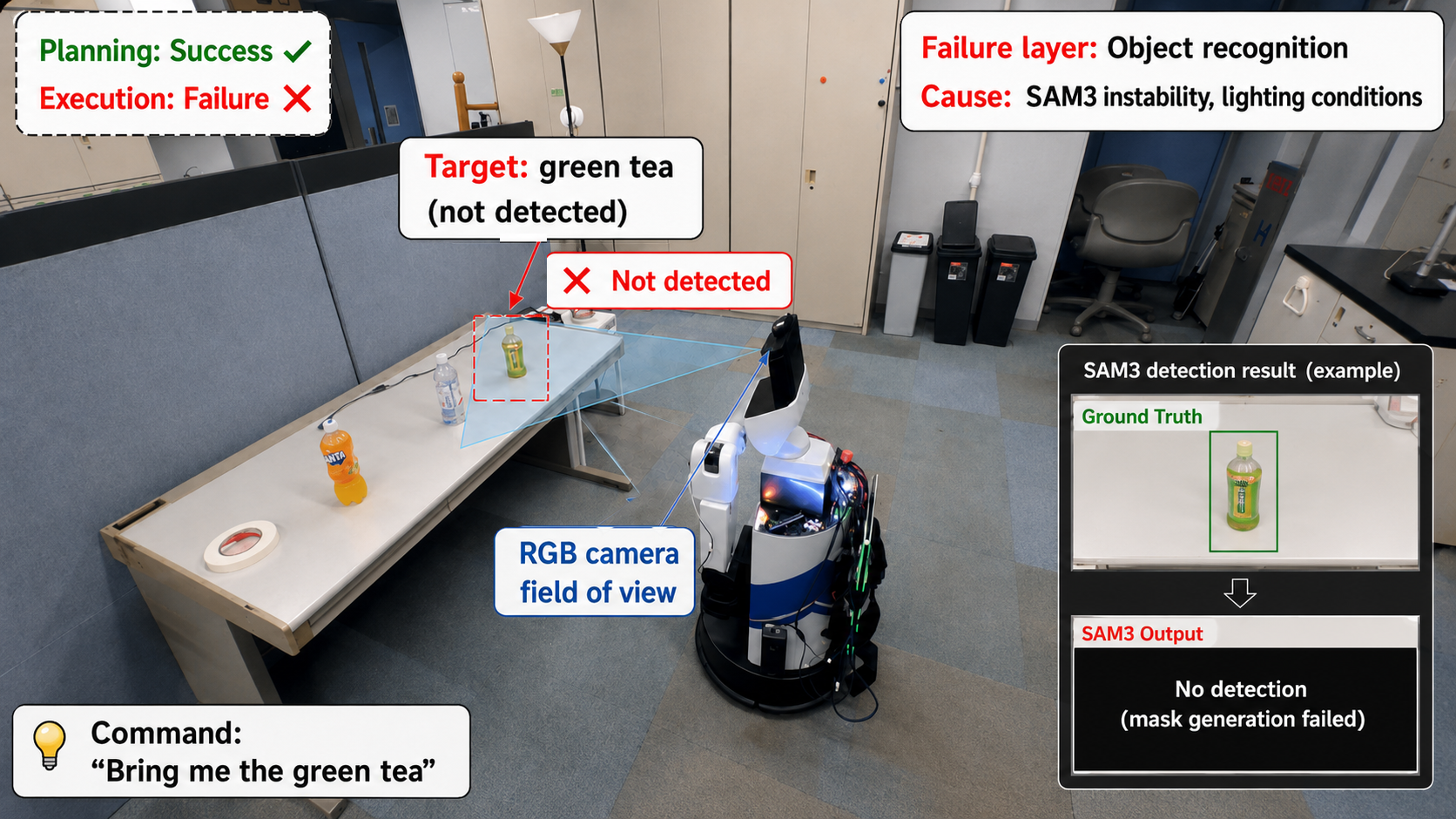}}
        \caption{Recognition failure: SAM3 returned no mask for the
        target object.}
        \label{fig:hsr_a}
    \end{subfigure}\hfill
    \begin{subfigure}{0.327\textwidth}
        \centering
        \fbox{\includegraphics[height=3.2cm]{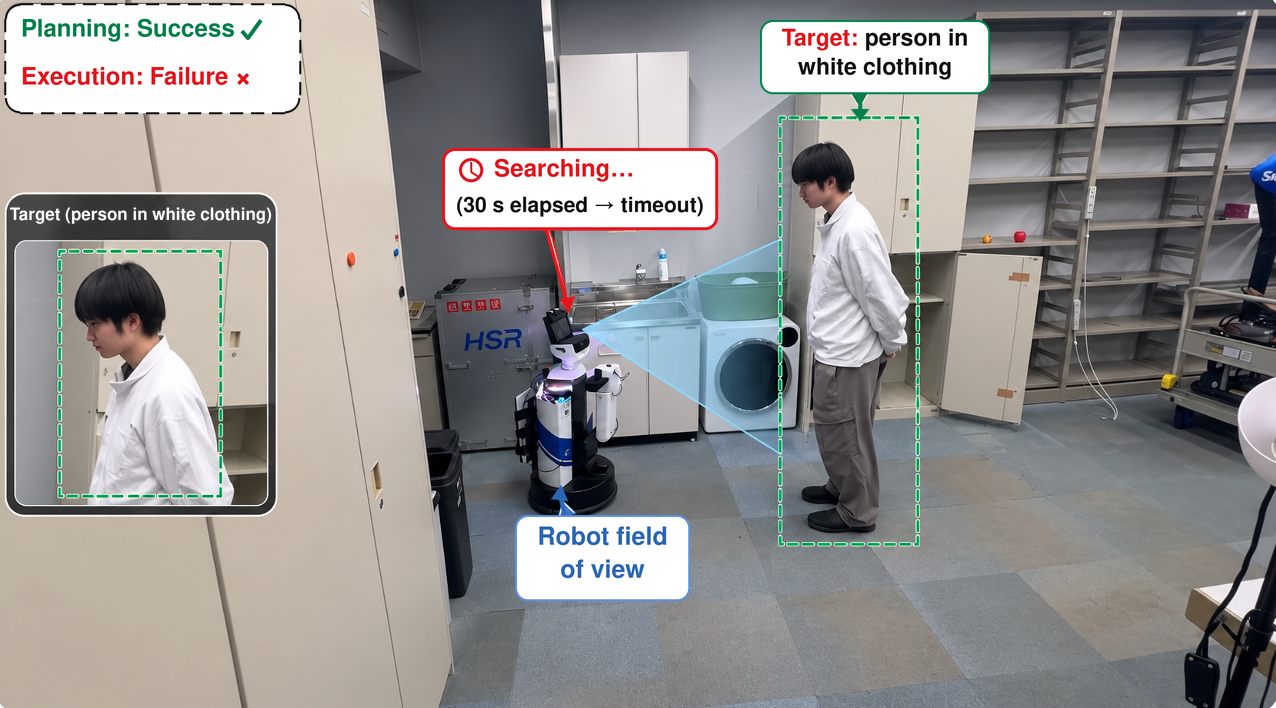}}
        \caption{Recognition failure: person identification exceeded
        the search window.}
        \label{fig:hsr_b}
    \end{subfigure}\hfill
    \begin{subfigure}{0.327\textwidth}
        \centering
        \fbox{\includegraphics[height=3.2cm]{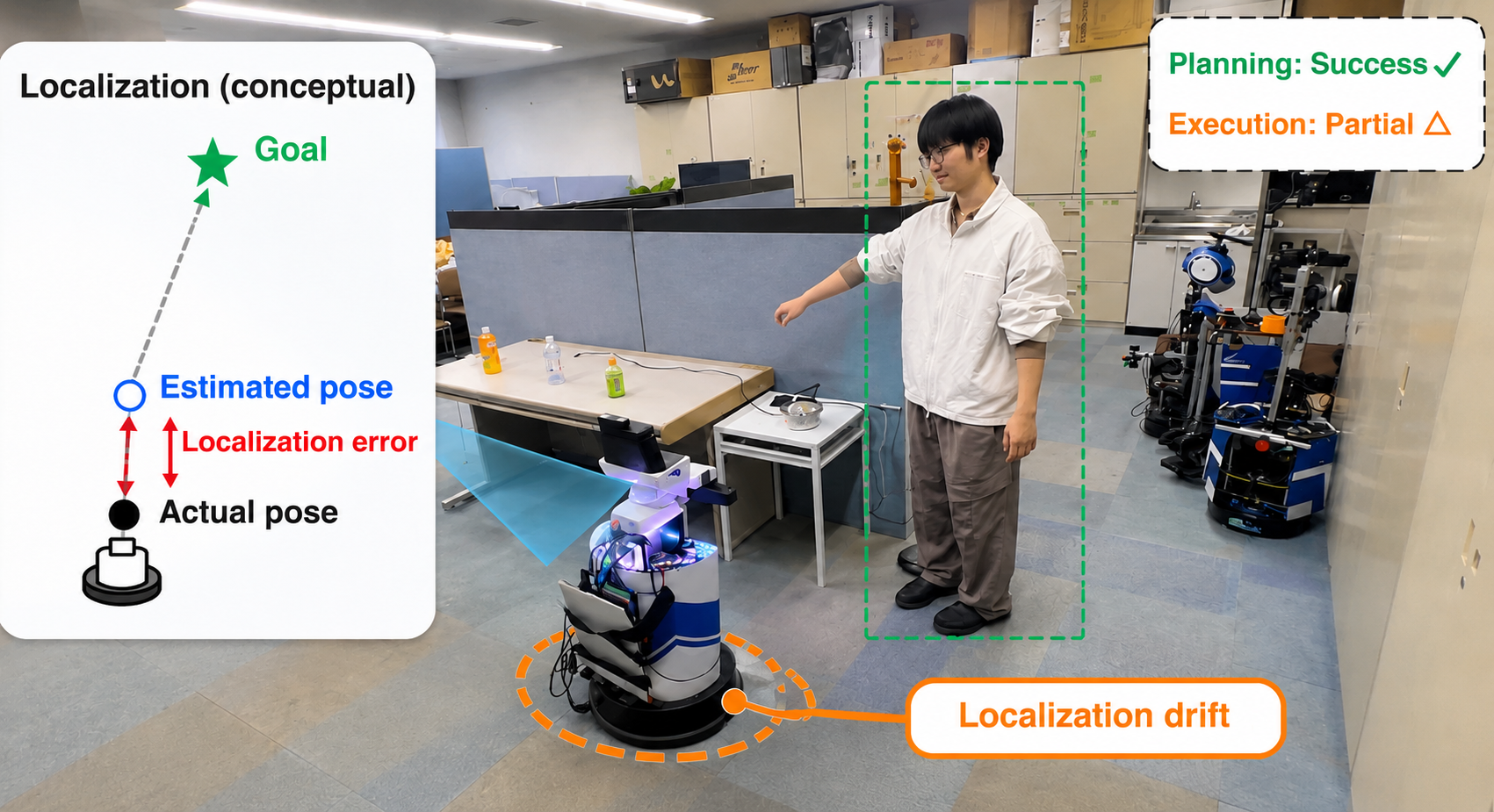}}
        \caption{Navigation failure: localization drift left the
        robot offset from the goal.}
        \label{fig:hsr_c}
    \end{subfigure}
    \caption{Three of the four execution-layer failures observed on
    the HSR. Planning succeeded in all three cases; the failures
    arose in the perception and localization components the plans
    depended on. See Table~\ref{tab:failures}.}
    \label{fig:hsr}
\end{figure*}

Planning succeeded in all 10 tasks; 6 tasks completed successfully
on the robot, while 4 did not fully complete due to execution-layer failures.
Table~\ref{tab:failures} summarizes three of these four cases, shown
in Fig.~\ref{fig:hsr}.

Fig.~\ref{fig:hsr}(a) shows a fetch task where SAM3~\cite{sam3}
returned no mask for the target bottle under ambient lighting, so
\texttt{DetectObject} returned empty and the sequence stalled before
grasping; the inset contrasts expected and returned segmentation.
In Fig.~\ref{fig:hsr}(b) the target person remained within the
camera field of view for the whole trial, yet feature-based
identification did not converge within the 30~s search window and
the task was abandoned at the \texttt{PersonFeature} step.
Fig.~\ref{fig:hsr}(c) shows the partial case: navigation reached the
neighbourhood of the goal, but accumulated localization error left
the robot offset from the intended pose, so the following
interaction step executed from a position the plan had not
anticipated.
In all three cases the generated sequence was valid; failure arose
from perception or localization.
Recognition failures were the most frequent, and no manipulation-layer
failures were observed in these 10 trials.

\begin{table}[!b]
\caption{Execution-Layer Failures on the HSR (three of four cases)}
\label{tab:failures}
\centering
\setlength{\tabcolsep}{3pt}
\renewcommand{\arraystretch}{1.25}
\small
\begin{tabular}{lccll}
\toprule
\textbf{Task} & \textbf{Plan} & \textbf{Exec}
              & \textbf{Layer} & \textbf{Cause} \\
\midrule
Fetch green tea
    & \checkmark & \texttimes
    & Recognition & SAM3 unstable \\
Guide by shirt color
    & \checkmark & \texttimes
    & Recognition & Timeout \\
Follow waving person
    & \checkmark & $\triangle$
    & Navigation & Localization drift \\
\bottomrule
\end{tabular}
\end{table}

\subsubsection{Discussion}
These results reveal a critical gap: even when LLM chaining produces
correct action sequences, execution-layer components --- object
recognition, person detection, and navigation --- introduce failures
that planning improvements alone cannot address, suggesting
planning benchmarks are insufficient to predict real-world
task completion.
The modular two-stage structure of the proposed architecture
naturally supports feedback-aware extensions as future work:
execution logs could be routed back to trigger targeted re-planning
at either the classification or action generation stage, enabling
self-recovery without modifying the core chaining design.
The failures in Fig.~\ref{fig:hsr} provide concrete triggers: an empty
detection or expired search window could prompt the Action Generator
to try another sub-location rather than terminate the task.

\subsection{Limitations}
The planning benchmark covers three models and a single command
source, so the reported gains may not transfer to other model
families or to command distributions outside the RoboCup@Home Japan
Command Generator.
Because the chaining condition used ground-truth task categories,
the reported gains are an upper bound under correct classification;
end-to-end performance would additionally depend on Stage~1
accuracy, which we have not measured on this benchmark.
The real-robot evaluation comprises 10 trials in one laboratory,
sufficient to identify failing execution layers but not estimate
failure rates reliably; the absence of manipulation failures may
reflect the small sample.
The 28 categories were derived by iterative refinement over
observed commands rather than by an independent procedure, so
coverage of unseen GPSR phrasings remains to be validated.

\section{Conclusion}
\label{sec:conclusion}
We proposed an LLM chaining architecture decomposing GPSR task
planning into instruction classification and action generation,
reducing per-inference context by $\approx$45\% and achieving
consistent SR improvements over SP across all models (up to
+37~pp on local models), together with longer sustained action
sequences on every model.
This reduction is per inference; total task-level token usage was not evaluated.
Real-robot experiments further identified execution-layer failures
as the primary remaining bottleneck.

Closing that gap is harder than improving planning alone because the planner cannot inspect perception or localization failures and therefore may repeat them under the same conditions.
The staged architecture naturally supports execution feedback,
associating outcomes with the task-specific schema that produced them.
Accumulated across trials, this would let a system build
category-level knowledge of which strategies hold under which
conditions --- for example, that a given object is unreliable to
segment under certain lighting and is better approached from a
different sub-location.
Combining a staged planner with the adaptivity of learned execution
policies is the most promising route toward GPSR
systems that improve from their own failures rather than repeat
them.

\let\oldthebibliography\thebibliography
\renewcommand{\thebibliography}[1]{%
  \oldthebibliography{#1}%
  \footnotesize
  \setlength{\itemsep}{0pt}%
  \setlength{\parskip}{0pt}%
}

\end{document}